# Joint Event Extraction along Shortest Dependency Paths using Graph Convolutional Networks

Ali Balali[a], Masoud Asadpour[a,*], Ricardo Campos[b,d], Adam Jatowt[c]

[a] *School of ECE, College of Engineering, University of Tehran, Tehran, Iran*

[b] *Ci2 - Smart Cities Research Center, Polytechnic Institute of Tomar, Tomar, Portugal*

[c] *Graduate School of Informatics, Kyoto University, Kyoto, Japan*

[d] *LIAAD - INESC TEC - INESC Technology and Science, Porto, Portugal*

**Abstract**

Event extraction (EE) is one of the core information extraction tasks, whose purpose is to automatically identify and extract information about incidents and their actors from texts. This may be beneficial to several domains such as knowledge bases, question answering, information retrieval and summarization tasks, to name a few. The problem of extracting event information from texts is longstanding and usually relies on elaborately designed lexical and syntactic features, which, however, take a large amount of human effort and lack generalization. More recently, deep neural network approaches have been adopted as a means to learn underlying features automatically. However, existing networks do not make full use of syntactic features, which play a fundamental role in capturing very long-range dependencies. Also, most approaches extract each argument of an event separately without considering associations between arguments which ultimately leads to low efficiency, especially in sentences with multiple events. To address the two above-referred problems, we propose a novel joint event extraction framework that aims to extract multiple event triggers and arguments simultaneously by introducing shortest dependency path (SDP) in the dependency graph. We do this by eliminating irrelevant words in the sentence, thus capturing long-range dependencies. Also, an attention-based graph convolutional network is proposed, to carry syntactically related information along the shortest paths between argument candidates that captures and aggregates the latent associations between arguments; a problem that has been overlooked by most of the literature. Our results show a substantial improvement over state-of-the-art methods.

*Keywords*: Information extraction, Event extraction, Deep learning, Shortest dependency path, Graph convolution network.

# 1 Introduction

Nowadays, a large amount of data is generated, especially, on the Internet. Social media and online news agencies play the main roles in producing this data. However, a vast majority of them is unstructured thus cannot be easily understood [1, 2]. Also, the large volume of available data makes it difficult for people to process it in a timely manner. One way to overcome this issue is to make use of Information Extraction (IE) as a means to automatically extract knowledge from unstructured data and give it a structured representation [3].

Event Extraction (EE) is among the most important sub-tasks of IE. Its aim is to automatically extract specific knowledge of certain incidents identified in texts [4] in the form of who is involved, in what, at when and where [5]. This task can be very beneficial in a variety of domains including question answering [6, 7], information retrieval [8], summarization [9-12], timeline extraction [13, 14], news recommendation [15, 16], knowledge base construction [9, 17], and online monitoring systems such as ones for health, life, disease, cyber-attack, stock markets, accident and robbery [18-24].

---

[*] This is to indicate the corresponding author.

Email address: balali.a67@ut.ac.ir (A. Balali), asadpour@ut.ac.ir (M. Asadpour), ricardo.campos@ipt.pt (R. Campos), jatowt@gmail.com (A. Jatowt)

As defined in Automatic Content Extraction (ACE) program[1], the problem of Event Extraction is, given a text, to identify event triggers with specific subtypes and their arguments for each sentence. To this regard, three subtasks have been defined: (1) Trigger Classification; (2) Argument Identification; (3) Argument Role Classification. Each one is described in more detail below.

The aim of **Trigger Classification** task is to identify event triggers and to classify them into their specific type (e.g., "*Life*") and corresponding subtype (e.g., "*Marry*"). According to ACE, an event trigger is the main word(s) that most clearly express the occurrence of an event. They serve to answer questions such as "*what happened*" by means of a verb (e.g. "*leaved*"), a noun (e.g. "*summit*"), and, occasionally through adjectives like "*former*". The ACE 2005 dataset includes 33 sub-types such as "*Marry*", "*Transport*", "*Attack*", etc., grouped into 8 event types ("*Life*", "*Movement*", "*Transaction*", "*Business*", "*Conflict*", "*Contact*", "*Personnel*" and "*Justice*"). The **Argument Identification** task, instead, aims to identify whether an entity mention, value or temporal expression is an argument for a particular event trigger or not. An entity mention (usually a noun phrase (NP), e.g., "*the largest Nations*", "*Bush*", "*France*"), a temporal expression (e.g., "*today*") or a value (e.g., "*Democratic National Chairman*") is a reference to an entity, an object or a set of objects in the world. According to ACE, an entity mention is limited to seven types: "*Person*", "*Organization*", "*Geo-political*", "*Location*", "*Facility*", "*Vehicle*" and "*Weapon*". Also, a value is limited to five types: "*Contact-Info*", "*Numeric*", "*Crime*", "*Job-Title*" and "*Sentence*". For example, the entity mention "*the largest Nations*" refers to the type "*Geo-political*", the temporal expression "*today*" refers to the type "*Time*" or the value "*Democratic National Chairman*" refers to the type "*Job-Title*". Finally, the **Argument Role Classification** task is to determine the role of an identified event argument. As referred by [25] in his research dated from 1980, this can be the general roles of time (*when*), location (*where*), source/actor (*who*: the initiator of the event), target (*whom*: the recipient of the event) and instruments (*how*: with what methods). Also, according to ACE, there are many other specific roles for arguments such as "*Buyer*", "*Artifact*", "*Victim*", "*Entity*", "*Destination*", etc.

Figure 1, shows an example of the event extraction task (in the lower part of the figure) and a dependency parser result (in the upper part) for the sentence "*Bush and Putin were leaved after their talks for the Group of Eight summit of the largest Nations in France*" taken from the ACE 2005 dataset.

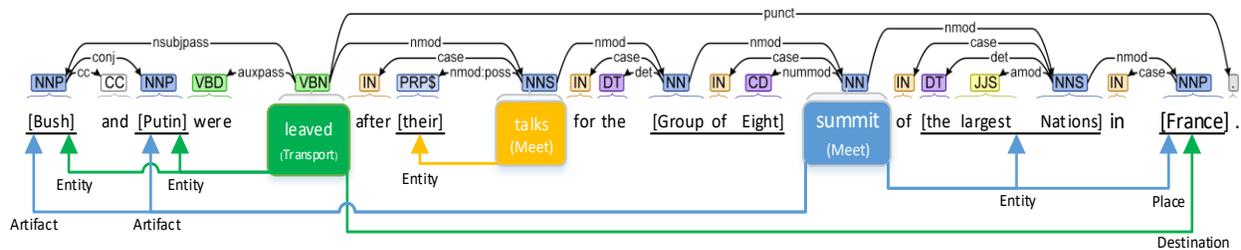

Figure 1: An example of the Event Extraction task (lower part of the figure) and its dependency parsing result (upper part). Source: the dependency parsing result is produced by Stanford CoreNLP[2] toolkit on a sentence of the ACE 2005 dataset.

In the figure, one can identify three **event triggers**: "*leaved*" (rectangular green box), "*talks*" (yellow box) and "*summit*" (blue box), and their corresponding subtypes: "*Transport*" (which belongs to the *movement* type), "*Meet*" (contact type) and "*Meet*", respectively. The **arguments** and corresponding **roles** associated with each event trigger can also be found in the sentence. For instance, "*Bush*", "*Putin*" and "*France*" (see green row) are arguments of the event trigger "*leaved*" with the roles "*Entity*", "*Entity*" and "*Destination*", respectively. Instead, "their" (see yellow row) is the argument of the event trigger "*talks*" with the role "*Entity*". Finally, "*Bush*", "*Putin*", "*the largest*

---

[1] https://catalog.ldc.upenn.edu/LDC2006T06
[2] http://nlp.stanford.edu:8080/corenlp/

*Nations*" and "*France*" (see blue row) are the identified arguments of the event trigger "*summit*" with the roles "*Artifact*", "*Artifact*", "*Entity*" and "*Place*", respectively. A careful analysis of this figure, also shows that while the "*Group of Eight*" has been identified as an entity mention (the same way in which "*Bush*" or "*Putin*" were) it was not assigned to any of the three event triggers ("*leaved*", "*talks*" or "*summit*"). From the figure, we can also observe a dependency parsing result, which describes the grammatical structure of the sentence, a very useful feature in the context of EE to understand the relationship among the words. For instance, knowing which are the subjects and the objects of a verb (e.g., an event trigger) would be extremely helpful to identify the arguments of an event trigger. In the concrete example of our figure, we can also observe that there is the dependency relation of "*nsubjpass*" between the argument "*Bush*" and the event trigger "*leaved*", from which we can induce a role to argument "*Bush*" in the event trigger "*leaved*".

Despite considerable advances in recent years, the area of EE still poses several challenges, most notably, (1) the ambiguity in classifying event triggers [26], (2) the error propagation problem among the sub-tasks of EE [12, 27, 28], (3) the existence of very long-range dependencies between event triggers and arguments [29] and (4) the lack of considering the associations between arguments [26, 30]. In the following, we describe each one of these challenges in more detail.

The complexity of natural language poses several problems in the EE task, including low efficiency that stems from the **ambiguity in classifying event triggers** in a sentence. For example, given the sentence "*He has fired his air defense chief*", event trigger "*fired*" could be classified as "*End-Position*" or "*Attack*" subtypes. However, if EE method knows that the context words "*air defense chief*" is a job title, the trigger type "*End-Position*" is selected [26]. Thus, to successfully identify an event trigger, the EE methods need to understand the context/semantics of words. Using neural layers such as the Long Short Term Memory (LSTM) [31] can help along this way. In recent years, some word embedding methods such as BERT[3] [32] and ELMO[4] [33] were introduced as a means to infer the context of a word in a sentence. These, as shown by [34] may be very helpful in solving ambiguity in classifying event triggers.

The **error propagation problem** occurs when event triggers and arguments are extracted in two stages as a pipeline approach [12, 27, 28]. In other words, the error propagates from the upstream classifier (the trigger classification task) to the downstream classifier (the argument identification task or the argument role classification task). To address this problem, methods for joint extraction of event triggers and arguments simultaneously have been proposed [28-30]. This approach also contributes to understanding how facts tie together in a sentence. For example, a "*Victim*" argument for a "*Die*" event is often the "*Target*" argument for the "*Attack*" event in a sentence explored by the joint approach [28].

The **very long-range dependencies** between event triggers and arguments -where information are close in context but distant in the sentence- is yet another problem in EE, which ultimately leads to low efficiency in sequential modeling methods [29]. This is evident in Figure 1, where we need 15 hops (walks) from the trigger candidate "*leaved*" to the argument candidate "*France*" if we follow a sequential order approach. As an attempt to solve this problem, a few works have focused on using syntactic information in the dependency graph during the design of neural architectures. This is the case of Liu et al. [29] who uses the graph convolution network (GCN) to capture the syntactic information flow from a point to its target through fewer transitions. Adopting such an approach would require only 5 hops instead of the previous 15 hops (along the *nmod* from "*leaved*" to "*talks*", from "*talks*" to "*Group*", from "*Group*" to "*summit*", from "*summit*" to "*Nations*", and from "*Nations*" to "*France*"). These five arcs in the dependency graph make the shortest dependency path (SDP) between the event trigger

---

[3] Bidirectional Encoder Representations from Transformers
[4] Embeddings from Language Models

"*leaved*" and the argument candidate "*France*" used in the proposed framework. Though an interesting solution, that work suffers from large number of learning parameters and high time complexity.

Finally, most of the previous approaches tend to identify different arguments of an event without considering **possible associations between arguments**, which may ultimately lead to low efficiency. For example, knowing that "*Bush*" and "*Putin*" are paralleled in the sentence[5], would ease the process of determining that they have the same role in the corresponding event triggers. To overcome this challenge, some research works use a Neural Tensor Network (NTN) to model the interaction among the arguments [26, 30]. However, tensor layers are very simple and cannot fully model the inter-dependencies between arguments.

In this work, we give particular attention to the main problems cited above. Specifically, we aim to tackle: 1) the very long-range dependencies problem; 2) the lack of considering the association between arguments; and 3) the ambiguity in classifying event triggers. Note however that, our proposed framework addresses the "error propagation problem", by joint extraction of event triggers and arguments recently discussed in the literature [28-30].

To tackle the above-referred problems, we propose a novel joint event extraction framework that aims to extract multiple event triggers and arguments simultaneously. In the *trigger classification task*, we use the concatenation of pre-trained Glove and Bert embedding in word representation layer, via a bidirectional Long Short Term Memory (LSTM) to capture the valuable semantics of a whole sentence automatically, as a means to solve the **ambiguity in classifying event triggers**. In the *argument identification* and *role classification tasks*, we calculate Shortest Dependency Paths (SDPs) in the dependency graph to extract syntactic information paths in sentences as a means to tackle the **very long-range dependencies problem**. This has been used by [35-38] to extract the relationship between two entities in the relation extraction task. Instead, in our paper, we use it to retain the most relevant information between the event trigger and the argument candidate by eliminating irrelevant words in-between the two. This, as shown before in the example of the event trigger "*leaved*" and the argument "*France*", would reduce the distance from 15 hops to just 5. Then, we use a Convolutional Neural Network (CNN) with dynamic multi pooling (DMCNN) [39] to automatically extract the most important features of the different parts of the SDP. Finally, a Graph Convolutional Network (GCN) via a self-attention layer is proposed to capture and aggregate the **associations between argument candidates** according to the length of shortest dependency paths (SDP-Ls). The GCN layer learns syntactic contextual representations of each node by the representative vectors of its immediate neighbors in the graph [40]. The proposed GCN layer encodes argument candidates as nodes and SDP-Ls as the label of edges. The SDP-L between argument candidates gathers useful information about the inter-dependency between argument candidates which can be captured by the proposed GCN layer. For example, the SDP-L=1 with the label "*conj*" between two-argument candidates (e.g., "*Bush*" and "*Putin*") shows a parallel structure that leads to the same role for them (e.g., "*Entity*"). Also, the SDP-L=2, typically shows two-argument candidates which have the same dependency parent. This is the case of the two-argument candidates "*Group of Eight*" and "*the largest Nations*" which have the same parent "*summit*" when considering the SDP-L=2. While using GCN layers may be an interesting solution to solve this problem, it involves many learning parameters as shown by Liu et al. [29]. For example, in Figure 1, six hops are needed to reach from the argument candidate "*Bush*" to the argument candidate "*France*" according to the shortest path on the dependency graph. Since one-layer GCN encodes only information about immediate neighbors, we need 6-layers of the GCN in order to consider the association between them. In contrast, in this work, we propose an improved version of the GCN layer by using SDP-Ls. Since the proposed GCN considers the inter-dependency between all argument candidates according to their SDP-L in one layer, we do not need to use a k-layer of them repeatedly to consider the long-range dependencies.

---

[5] By means of their connection through the "and" word.

The main contributions of this work are as follows:

- We propose a novel event extraction framework based on deep neural networks, named *JEE-SDP*, to predict event triggers and their arguments. The proposed framework achieves 75.83% F1-score in the trigger classification task (2.1% improvement compared to deep neural networks state-of-the-art methods) and 66.44% F1-score in the argument role classification task (3.69% improvement) on the popular ACE benchmark dataset;
- For the *trigger classification task*, we focus on solving the ambiguity problem in classifying event triggers. For this purpose, we use the concatenation of pre-trained Glove and Bert embedding in word representation layer, via a bidirectional Long Short Term Memory (LSTM) to improve the representation of words in a sentence by considering their context;
- For the *argument identification and role classification tasks*, we focus on solving the very long-range dependencies problem and the associations between argument candidates. For the former (very long-range dependencies problem between event triggers and argument candidates), we calculate Shortest Dependency Paths (SDPs) in the dependency graph to extract the most relevant information between the event trigger and the argument candidate by eliminating irrelevant words in-between the two. For the latter (associations between argument candidates), we capture and aggregate the inter-dependency between argument candidates according to syntactic information path (based on the length of the SDP between them) by applying a GCN layer via a self-attention layer.

The remainder of this paper is structured as follows. Section 2 describes the related work. Section 3 describes our *JEE-SDP* framework in detail. Section 4 introduces the experimental part of the work and describes the results. Finally, Section 5 concludes the paper.

## 2 Related Work

One of the first attempts towards solving the problem of EE was conducted in 1960s by DARPA on WEIS[6] project [41]. Subsequently, other EE projects emerged, most notably, the COPDAB[7] [25], MID[8] [42], IDEA[9] [43], ACE[10] [44], CAMEO[11] [45], TAC[12] KBP [46], ICEWS[13] [47] and PLOVER[14]. Ultimately, the majority of the activities begun in 2004 boosted by the ACE program, which has been used to a large extent by many researchers to construct comprehensive models [26, 48-50].

Despite significant advances in the last few years, the problem of extracting events from text with high efficiency still poses several challenges, such as ambiguity in classifying event triggers [26], the lack of enough training data [12, 34], the roles overlap problem [34], the error propagation problem among the sub-tasks of EE [12, 28, 51] and the existence of very long-range dependencies between event triggers and arguments [29]. Most of the research conducted so far, has tried to tackle these problems by employing (1) feature-based; (2) deep neural network-based; or (3) data augmentation-based methods. Following, we briefly summarize some of the most important works in this respect. A more detailed description of those selected as baselines will be given in the experiments section.

---

[6] World Event/Interaction Survey
[7] The Conflict and Peace Data Bank
[8] Militarized Interstate Disputes, http://cow.dss.ucdavis.edu/data-sets/MIDs
[9] Integrated Data for Events Analysis
[10] Automatic Content Extraction
[11] Conflict and Mediation Event Observations
[12] Text Analysis Conference
[13] Integrated Crisis Early Warning System.
[14] Political Language Ontology for Verifiable Event Records, https://github.com/openeventdata/PLOVER

*Feature-based methods* focus on designing various hand-crafted feature sets such as lexical features (e.g., POS-tag) and syntactic features (e.g. dependency relations) to extract events. For instance, Ahn [52] considered a set of WordNet and dependency features to identify events. Li et al. [27] proposed some global features to capture the dependencies of multiple event triggers and arguments. Finally, some works [53, 54] proposed the cross-event and cross-entity features to capture the relation of events and arguments. Although extensively used in the early days of EE, selecting a suitable feature set turned into a problem in feature-based methods due to requiring extensive human engineering. **Deep neural networks-based methods**, instead, make use of neural network layers, such as CNN with dynamic multi pooling [39], LSTM with dependency bridges [30] and Bi-SRU[15] based on words and characters [55] as a means to extract underlying features in sentences automatically. While considered a valid solution, they suffer from low efficiency in capturing very long-range dependencies [29]. Recent years brought into play some works focusing on solving this problem by using syntactic information during the design of neural architectures [29]. For instance, Nguyen et al. [28] used a binary memory vector to model dependency relations [28], yet the method is still weak in terms of syntactic modeling. Liu et al. [29] used graph convolution networks (GCNs) to model syntactic information in the dependency graph. GCNs learn syntactic contextual representations of each node by the representative vectors of its immediate neighbors in the graph. One layer GCN encodes only information about immediate neighbors and to encode *K*-order neighborhoods, *K* layers are needed [56]. GCN layers are complementary to LSTM layers, when both GCN and LSTM layers stack together [56]. However, GCNs suffer from the large number of learning parameters. Thus, the effectiveness of GCNs is limited in tasks which have small training data like EE. Also, using *K* layers of the GCN needs many computations which is time-consuming. To model the interaction among words or arguments, some works [26, 30, 57] used a neural tensor network (NTN) which was found to be effective for capturing multiple interactions among words [57] and matching text sequences to compare aggregate models [58]. Despite these aspects, tensor layers are very simple and cannot fully model the associations among arguments. Due to the large number of learning parameters in complex neural network layers, they especially suffer from small training data. Finally, **Data augmentation-based methods** tackle the problem of data scarcity by using extra knowledge. These methods use augmentation techniques to enhance the training data using other datasets such as FrameNet [12, 59], FreeBase [60] or embedding methods such as BERT [34]. Also, some other works [5, 61] make use of tools that allow users to find, expand and annotate event triggers by exploring unannotated data.

## 3 JEE-SDP Approach

In this section, we describe *JEE-SDP,* a joint approach framework whose aim is to simultaneously extract event triggers and arguments for a sentence. Figure 2, illustrates the architecture of *JEE-SDP* for our running example: "*Bush and Putin were leaved after their talks for the Group of Eight summit of the largest Nations in France*". Our framework involves three tasks: (1) Trigger classification; (2) Argument identification; and (3) Argument role classification. Each of these tasks consists of several modules. For the **trigger classification task**, we have three modules: Embedding layer (Section 3.1.1); Bi-LSTM layer (Section 3.1.2); and Trigger classification output (Section 3.1.3). As the **argument identification** and the **argument role classification tasks** have the same architecture in *JEE-SDP* (except in the last layer, i.e., the output of the model), we present them together for a matter of simplicity. Both make use of three modules: SDP-DMCNN (Section 3.2.1); GCN-SDP (Section 3.2.2); and Output layer (Section 3.2.3). The tasks and their corresponding modules will be described in-depth in the following sub-sections.

---

[15] Bi-directional simple recurrent unit

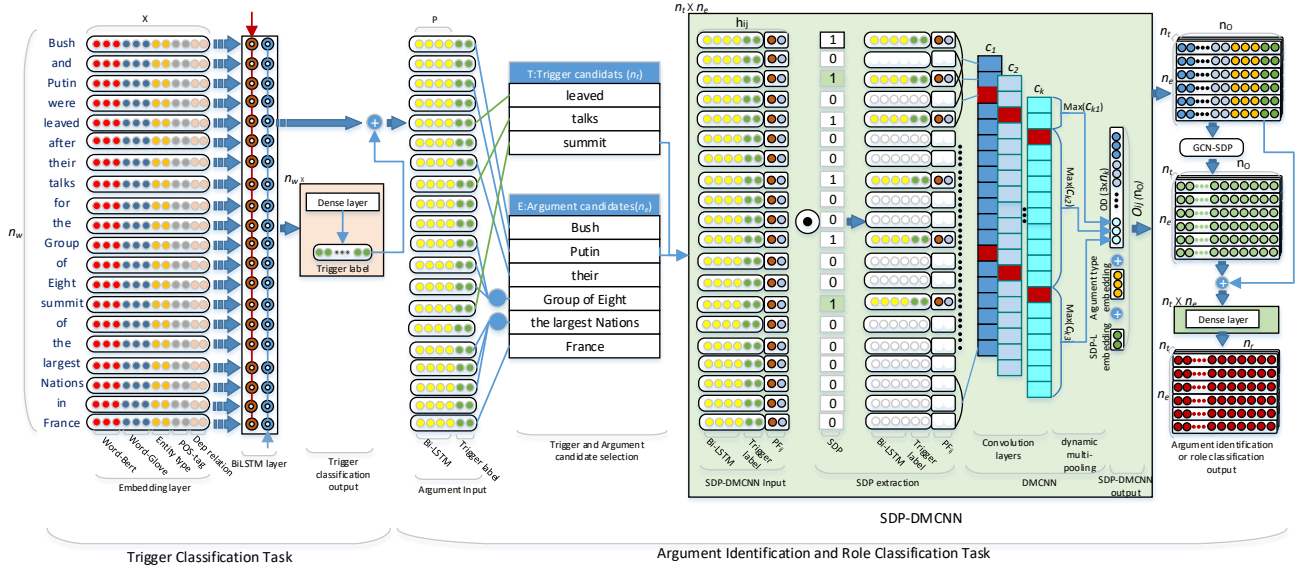

Figure 2: Architecture of *JEE-SDP* for the event extraction task depicted on our running example sentence taken from the ACE 2005 dataset. In the SDP-DMCNN module, the processing of the event trigger "*summit*" and the argument candidate "*Putin*" is illustrated.

### 3.1 Trigger Classification Task

In this section, we describe the trigger classification task in more detail. Given a sentence $W = \{w_1, w_2, ..., w_{n_w}\}$ with length $n_w$, each token $w$ is represented by a real-valued vector $x$ that results from embedding layer (Section 3.1.1). Each of these vectors is then passed into a Bi-LSTM layer (Section 3.1.2) to consider the semantics of the words. Finally, the output layer predicts a trigger label for each token (Section 3.1.3).

### 3.1.1 Embedding Layer

In our framework, each input token $w$ of a sentence is transformed to a real-valued vector $x$ which stems from concatenating Bert, Glove, entity type, POS-tag and dependency relation embeddings. In the following, we describe each one of them in more detail.

- **Word Embedding**

In our work, we make use of two well-known pre-trained word embedding methods, Glove[16] [62] and Bert[17] [32], to create a vector representation of the token $w_i$ learned by understanding the context of words. Glove is a context-free model which generates a single word embedding representation for each token $w_i$ of the vocabulary, by simply looking up for the corresponding entry in the pre-trained word vectors. BERT, on the other hand, does this by jointly conditioning on both left and right contexts. In this research paper, we use "*BERT-Base, Uncased*" version of BERT, from a Github repository[18]. Using both methods together is an important contribution to understand the context of a token $w_i$.

- **Entity Type Embedding**

---

[16] Global Vectors for Word Representation, "https://nlp.stanford.edu/projects/glove/"
[17] Bidirectional Encoder Representations from Transformers
[18] https://github.com/google-research/bert

In our work, entity mentions are annotated with the BIO annotation schema. The following sentence "*former Democratic National Chairman Ron Brown*" illustrates the identification of 5 entity mentions and their entity types (Person, Time, Organization and Job-Title):

- Person: "*former Democratic National Chairman Ron Brown*"
- Person: "*former Democratic National Chairman*"
- Time: "*former*"
- Organization: "*Democratic*"
- Job-Title: "*Democratic National Chairman*"

This example highlights the fact that each token $w_i$, refers to different entity types. For instance, the token "*Democratic*" can be found in three different entity types ("*Person*", "*Organization*" and "*Job-Title*"). This problem, require us to represent each token $w_i$, as the sum of all different entity types embedding vectors in which the token participates. For example, to encode the word "*Democratic*", we need to sum the three entity type embedding vectors "*Person*", "*Organization*" and "*Job-Title*". Each entity type vector is represented as a one-hot vector encoding, that is, a vector of 0s and a single 1, whose length is the number of different entity types.

- **POS-Tag Embedding**

A unique vector is randomly generated according to the size of the POS-tag embedding vector and then assigned to each POS-tag.

- **Dependency Relation Embedding**

The dependency relation of each token $w_i$ shows syntactic relation of $w_i$ to its parent. Capturing this information, is possible by using some popular dependency parsers such as Stanford CoreNLP[19] [63] and Spacy[20]. In this paper, we use Spacy, currently the fastest NLP parser, which is also offering the best state-of-the-art accuracy [64]. According to the size of the dependency relation embedding vector, a unique vector is randomly generated and assigned to each dependency relation.

The 5 embedding vectors are then concatenated giving rise to a matrix $X \in R^{n_w \times n_d}$ where $n_w$ is the length of the sentence, and $n_d$ is the sum of the different embedding vectors size. The matrix $X$ is then fed into the Bidirectional LSTM layer.

### 3.1.2 Bidirectional LSTM (Bi-LSTM)

Given the matrix $X$, we use the Long Short Term Memory (LSTM) [31] with bidirectionality [65] to create vectors which are learned by understanding the context of words in forward and backward ways in the sentence. The output of the *i*-th token of a sentence in this layer is given by $P_i = [\vec{p}_i, \overleftarrow{p}_i]$, where $\vec{p}_i$ and $\overleftarrow{p}_i$ are, respectively, the outputs of the forward and backward LSTMs [29, 30] as represented by:

$$\vec{p}_i = LSTM^\rightarrow(\vec{p}_{i-1}, x_i)$$
$$\overleftarrow{p}_i = LSTM^\leftarrow(\overleftarrow{p}_{i+1}, x_i) \quad (1)$$

Also, the word embedding vectors which were obtained in the previous step, especially ones by BERT, need to be customized according to the event extraction task and the ACE dataset. Bidirectional LSTM can solve this problem by its hidden states which are learned during backpropagation over time.

### 3.1.3 Trigger Classification Output

---

[19] https://corenlp.run/
[20] https://spacy.io/

The outputs of the Bi-LSTM layer are then fed into a fully-connected network as follows:

$$O_i = f(W_1 P_i + b_1)$$
$$y_i = softmax(W_2 O_i + b_2) \qquad (2)$$

where, $1 \leqslant i \leqslant n_w$, and $W_1, W_2$ are the weight matrices, $b_1$ and $b_2$ are the bias of the gate, $f$ is a non-linear activation, and $y_i$ is the score of different trigger labels for token $w_i$ in the BIO annotation schema. Thus, the length of the trigger classification output for a token $w_i$ is equal to $2L+1$ where $L$ is the size of event trigger subtypes with the "*NoEvent*" label.

## 3.2 Argument Identification and Role Classification Task

In this section, we describe the architecture of *JEE-SDP* in the argument identification and role classification tasks. The input for both results from the concatenation of the vector $P$ (generated by the Bi-LSTM) and the output vector of the trigger classification task (which gives the trigger labels for each token $w_i$). Given a list of entity mentions[21] $E = \{e_1, e_2, ..., e_{n_e}\}$ as argument candidates (e.g., "*Bush*", "*Putin*", "*their*", "*Group of Eight*", "*The largest Nations*" and "*France*"), where $n_e$ is the number of entity mentions, and a list of trigger candidates $T = \{t_1, t_2, ..., t_{n_t}\}$ extracted from the previous layer (e.g., "*leaved*", "*talks*", "*summit*"), where $n_t$ is the number of trigger candidates, the SDP-DMCNN module aims to calculate the shortest dependency path (SDP) between a trigger candidate $t_i$ and an argument candidate $e_j$ (Section 3.2.1). Then, in order to capture and aggregate the associations between argument candidates $E$, the GCN-SDP module is applied (Section 3.2.2). Finally, the output layer predicts the role of each argument candidate $e_j$ for each trigger candidate $t_i$ (Section 3.2.3). In the following, we introduce each one of these three modules in more detail.

### 3.2.1 SDP-DMCNN

The SDP-DMCNN module is composed of four layers: the inputs, the Shortest Dependency Path (SDP) extraction and the Dynamic Multi-pooling Convolutional Neural Network (DMCNN) [39] and finally, the output layer. This module is executed for each trigger candidate $t_i$ and argument candidate $e_j$. Thus, for each execution, we select one trigger $t_i$ and one argument candidate $e_j$. Figure 2 shows the execution of the module for trigger candidate "*summit*" and argument candidate "*Putin*".

- **SDP-DMCNN Input**

The input of the SDP-DMCNN for the *k-th* token of a sentence would be a vector $h_{ijk} \in R^{n_h}$, with length $n_h$ for the trigger candidate $t_i$ and the argument candidate $e_j$ that results from the concatenation of the output of the Bi-LSTM layer ($p_k$), the output of the trigger classification task ($y_k$) and the positional embedding ($PF_k$):

$$h_{ijk} = p_k \oplus y_k \oplus PF_{ik} \oplus PF_{jk} \qquad (3)$$

Here, $PF_{ik}$ and $PF_{jk}$ specify the relative distance between a *k-th* token of a given sentence, with the trigger candidates $t_i$ and the argument candidate $e_j$ respectively. $PF$ is necessary to specify which words are the trigger and argument candidate [28, 39, 66]. To encode the position feature (*PF*), each distance value is represented by an embedding vector which is generated randomly. For example, in Figure 1, the relative distances of token "*Putin*" and the trigger candidate "*summit*" is 11. Also, the relative distance of token "*Putin*" and the argument candidate "*Bush*" is 2. Also note that $p_k$ and $y_k$ are already known from the previous steps.

In the following, we feed the input vectors ($h_{ij}$) to a SDP extraction layer to eliminate the irrelevant input vectors with considering the SDP of the trigger candidate $t_i$ and the argument candidate $e_j$.

---

[21] In order to simplify, entities, values and temporal expressions are presented as entity mentions in the rest of the paper.

- **SDP Extraction**

Using the shortest path between two entities in the dependency graph has proven to be a valid and helpful solution within the context of the relation extraction task [35-37]. In this research work, we apply the SDP to predict the role of the argument candidates. To accomplish this in the SDP-DMCNN module, we only consider the words which are along the SDP between the trigger candidate $t_i$ and the argument candidate $e_j$. For instance, we would consider the words "*Putin*", "*Bush*", "*leaved*", "*talks*", "*Group*" and "*summit*" when aiming to determine the SDP between "*Putin*" and "*summit*" in Figure 1. Also, the length of the SDP (SDP-L) between the trigger candidate $t_i$ and the argument candidate $e_j$ is important as it presents significant information about relationships.

---

**Algorithm 1:** The SDP calculation

**In**: A: dependency arc matrix ($\in \mathbf{R}^{n_w \times n_w}$), T: the position of trigger candidates ($\in \mathbf{R}^{n_t \times n_w}$), E: the position of argument candidates ($\in \mathbf{R}^{n_e \times n_w}$)
**Out**: SDP $\in \mathbf{R}^{n_t \times n_e \times n_w}$ & SDP-L $\in \mathbf{R}^{n_t \times n_e}$
**Initiate**: M: SDP between words ($\in \mathbf{R}^{n_w \times n_w \times n_w}$) $\leftarrow$ 0, ML: SDP Length matrix between words ($\in \mathbf{R}^{n_w \times n_w}$) $\leftarrow n_w$, A: dependency arc matrix ($\in \mathbf{R}^{n_w \times n_w}$) $\leftarrow$ 0, SDP-L ($\in \mathbf{R}^{n_t \times n_e}$) $\leftarrow n_w$, SDP ($\in \mathbf{R}^{n_t \times n_e \times n_w}$) $\leftarrow$ 0
1. **for** i $\leftarrow$ 0 to $n_w$ **do**
2.    M[i], ML[i] $\leftarrow$ BFS (i,A)
3. **end for**
4. **for** i $\leftarrow$ 0 to $n_t$ **do**
5.    V $\leftarrow$ T[i]
6.    **for** j $\leftarrow$ 0 to $n_e$ **do**
7.      U $\leftarrow$ E[j]
8.      **if** overlap (V,U) == false **then**
9.         **for** k $\leftarrow$ 0 to $n_w$ **do**
10.           **if** V[k] ==1 **then**
11.             **for** z $\leftarrow$ 0 to $n_w$ **do**
12.                **if** U[z] == 1 **then**
13.                   **if** SDP-L[i,j] > ML[k,z] **then**
14.                      SDP-L[i,j] $\leftarrow$ ML[k,z]
15.                      SDP[i,j] $\leftarrow$ M[k,z]
16.                  **end if**
17.                **end if**
18.             **end for**
19.           **end if**
20.         **end for**
21.      **end if**
22.      SDP[i,j] $\leftarrow$ SDP[i,j]+ U
23.    **end for**
24.    SDP[i,j] $\leftarrow$ SDP[i,j]+ V
25 **end for**

---

To calculate the SDP and the SDP-L between trigger and argument candidates we resort to Algorithm 1. In this algorithm, $A \in \mathrm{R}^{n_w \times n_w}$, is an adjacency matrix based on dependency arcs which are extracted from the dependency parser. We assume that arcs in the dependency graph are undirected. $T \in \mathrm{R}^{n_t \times n_w}$ and $E \in \mathrm{R}^{n_e \times n_w}$ are the matrices which indicate positions of trigger and argument candidates in a sentence. For example, the vector $e_4 \in \mathrm{R}^{n_w}$ in Figure 2, shows the position of the argument candidate *"The largest Nations"* in the sentence. The values of this vector are equal to zero, except in the indexes of 15, 16 and 17 which are equal to 1. At the first step (Lines 1-3), the path and distance between words in the dependency graph are calculated by the breadth-first search (BFS) algorithm [67]. Given a token $w_i$, BFS starts at token $w_i$ and then explores all of the neighbor tokens at the present depth prior to moving on to the tokens at the next depth level. The shortest paths between all words and their lengths are calculated by running the BFS on all tokens in the sentence. Since, trigger and argument candidates can include one or more tokens, the shortest path is calculated between all tokens of the trigger candidate $t_i$ and all tokens of the argument candidate $e_j$ and then, the minimum path length between them is selected as the SDP (Lines 4-21). Finally, we sum the SDP vector with two vectors $V_i$ and $U_j$ which show the positions of the trigger candidate $t_i$ and the argument candidate $e_j$, respectively (Lines 22, 24) in the sentence. In order to eliminate noise[22], the SDP calculation is ignored when there is a common token between the trigger candidate $t_i$ and the argument candidate $e_j$ (Line 8). Outputs of this algorithm are the SDP $\in R^{n_t \times n_e \times n_w}$ and the SDP-L $\in R^{n_t \times n_e}$. The SDP for the trigger candidate $t_i$ and the argument candidate $e_j$ presents a vector of 0s and 1s, which 1s show the index of tokens which are along the SDP. In Figure 2, the SDP vector is presented for the trigger candidate "*summit*" and the argument candidate "*Putin*".

To select tokens which are along the SDP, we use the element-wise product operation as follows:

$$h_{ij} - SDP = h_{ij} \odot SDP \qquad (4)$$

where, $h_{ij}$-SDP $\in R^{n_w \times n_h}$ includes the tokens which are along the SDP between the trigger candidate $t_i$ and the argument candidate $e_j$. $h_{ij}$-SDP is passed to the DMCNN layer described in the next section.

- **DMCNN**

Next, we make use of a convolution neural network (CNN) layer to extract informative features on SDPs. We do this, by defining some feature maps. Since, one sentence may contain two or more event triggers, and one argument candidate may play different roles in a sentence, a max-pooling layer on the whole sentence is not enough for EE. Thus, to extract the most important features after the convolution operation, a dynamic multi-pooling layer is used to get max value for each part of a feature map [39]. Feature maps are splitted to three parts by the position of trigger candidate $t_i$ and argument candidate $e_j$.

In the following, we describe DMCNN of Chen et al. [39][23]. A convolution operation defined $n_k$ filters $w \in R^{m \times n_h}$, where $m$ is the window size and $n_h$ is the length of input vector, which is applied to a window of $m$ tokens to produce new features. In Equation 5, a feature $c_{ij}$ is generated from a windows of tokens $x_{i:i+m-1}$ based on the filter $w_j$. This filter is applied to each possible window of tokens in the sentence.

$$c_{ij} = f(W_j \, \mathrm{x}_{i:i+m-1} + b_j) \qquad (5)$$

Where $1 \leqslant j \leqslant n_k$, $f$ is a non-linear function and $b$ is a bias. The size of $i$ is equal to the length of the sentence by padding special symbol. According to Figure 2, the feature map output $C_k$ is divided into three sections $C_{k1}$, $C_{k2}$ and $C_{k3}$ by "*Putin*" and "*summit*". In Equation 6, the dynamic multi-pooling is presented, where $1 \leqslant j \leqslant n_k$ and $1 \leqslant i \leqslant 3$:

$$od_{ji} = \max(C_{ji}) \qquad (6)$$

The $od_{ji}$ is calculated for each feature map and then, they are concatenated to form a vector $OD \in R^{3n_k}$.

- **SDP-DMCNN Output**

The output of the SDP-DMCNN$_{ij}$ $\in R^{n_o}$ for trigger candidate $t_i$ and argument candidate $e_j$ is obtained by:

$$O_{SDP-DMCNN_{ij}} = OD_{ij} \oplus \text{SDP} - L_{ij} \oplus \text{Argument} - \text{type}_j \qquad (7)$$

where, $\oplus$ is the concatenation operation and $OD_{ij}$ is the output of the DMCNN layer for trigger candidate $t_i$ and argument candidate $e_j$. The SDP-L and the argument type embeddings are explained in more detail as follows. **The shortest dependency path length embedding (SDP-L embeddings)** feature shows the length of the SDP between the trigger candidate $t_i$ and the argument candidate $e_j$ in the dependency graph which is calculated according to

---

[22] Typically, the SDP-L is equal to 1 when trigger and argument candidate has a common token.
[23] Please refer to Chen et al. (2015) for a comprehensive overview of DMCNN.

Algorithm 1. For example, in Figure 1, the SDP-L between "*Putin*" and "*summit*" is 5. Encoding this feature is a similar process to the one conducted with PF. **The argument type embedding,** encodes the type of entity mentions for each argument candidate $e_j$. We know that each argument candidate is an entity mention which has a specific type such as *person*, *location*, *vehicle, time* and etc. in a sentence. This argument type is a critical feature in predicting the role of argument candidates for each trigger candidate in a sentence. For example, an argument candidate with type *location* typically plays a few roles such as *Destination* and *Place* for different trigger candidates. To encode this feature, a unique vector is randomly generated and assigned to each argument type.

Once this process is concluded, the output of the SDP-DMCNN $\in R^{n_t \times n_e \times n_o}$ is passed to the GCN-SDP to consider the associations between argument candidates.

### 3.2.2 GCN-SDP

The commonality, differences and interactions between the argument candidates can be explored by deep neural network layers. Some types of interactions are, such as whether two argument candidates have the same dependency parent, whether two argument candidates are parallel in a sentences or whether two argument candidates are semantically coherent [30].

In this paper, the interactions between argument candidates are modeled by a graph convolution network (GCN) according to syntactic information paths. GCNs are neural networks operating on graphs and extracting features of nodes based on the properties of their neighborhoods [40, 56]. Typically, GCNs are used as encoding sentences which produce latent feature representations of each word according to syntactic arcs in the dependency graph. In this work, instead of using syntactic arcs in the dependency graph, the SDP-L among argument candidates has been used in the GCN (GCN-SDP).

Given a undirected graph G=($V$, $\mathcal{E}$), $V = \{v_1, v_2, \dots, v_{n_e}\}$ are the set of nodes which represent argument candidates and $\mathcal{E}$ is the set of edges between argument candidates. The label of each edge $(v_i, v_j) \in \mathcal{E}$ is the value of the SDP-L between two nodes. For example, in Figure 1, the label of the edge between two nodes *"Bush"* and *"France"* is 6 (SDP-L=*6*) which means that 6 hops are needed to reach *v("France")* from *v("Bush")* according to arcs in the dependency graph. For each SDP-L=*d* and argument candidate $e_i$, the GCN-SDP presents a weighted aggregation over some argument candidates which have the same SDP-L=*d* with $e_i$. Then, we use a self-attention layer to aggregate different SDP-Ls to prepare a vector for the argument candidate $e_i$. This vector captures a global view of the associations between $e_i$ with other argument candidates. In the following, we introduce two steps of the GCN-SDP in more detail.

- **SDP-L Adjacency Matrices Calculation**

In this step, we need to calculate a SDP-L matrix which shows SDP-Ls among argument candidates. The SDP-L matrix can be calculated according to Algorithm 1 (with the exception that SDP-Ls are calculated between argument candidates). The SDP-L between the same argument is assumed to be zero. Figure 3 shows the SDP-L matrix, where $n_e$ is the number of argument candidates and $n_s$ is the maximum length of the SDP between argument candidates. In this work, we assume that $n_s$=10. It means that we ignore SDPs whose length are more than 10. To feed SDP-L matrix to neural network layers, the SDP-L adjacency matrices are decomposed from the SDP-L matrix according to Figure 3. Each SDP-L adjacency matrix shows the relation between argument candidates in a specific SDP-L.

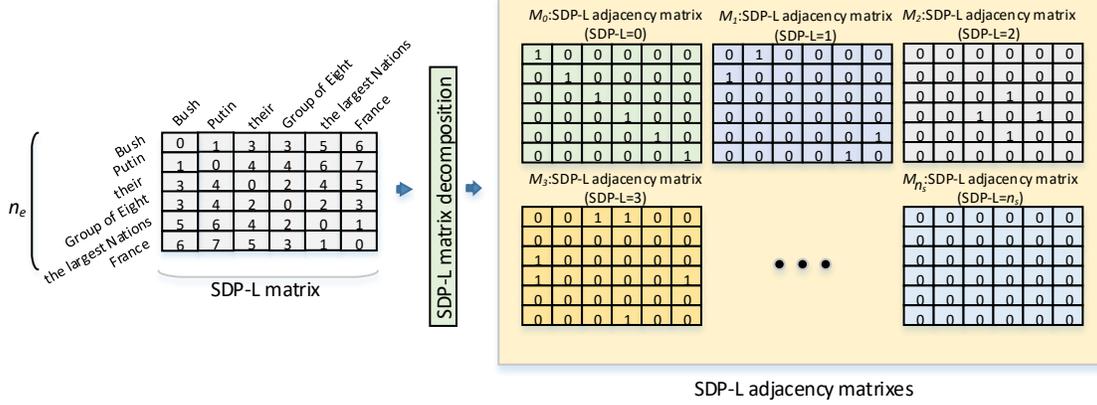

Figure 3: A example for preparation of SDP-L adjacency matrices between argument candidates.

- **Graph Convolution Network + Self-Attention Layer**

The GCN-SDP is customized in comparison to the GCN layers in the previous works [29, 40]. The GCN-SDP uses fewer learning parameters due to small training data in EE. Moreover, instead of using sum layer to aggregate $n_s$ different SDP-L vectors, a self-attention layer is used. The self-attention layer considers the importance of the different SDP-L vectors through learning parameters in aggregation phase. Figure 4, shows the GCN-SDP architecture.

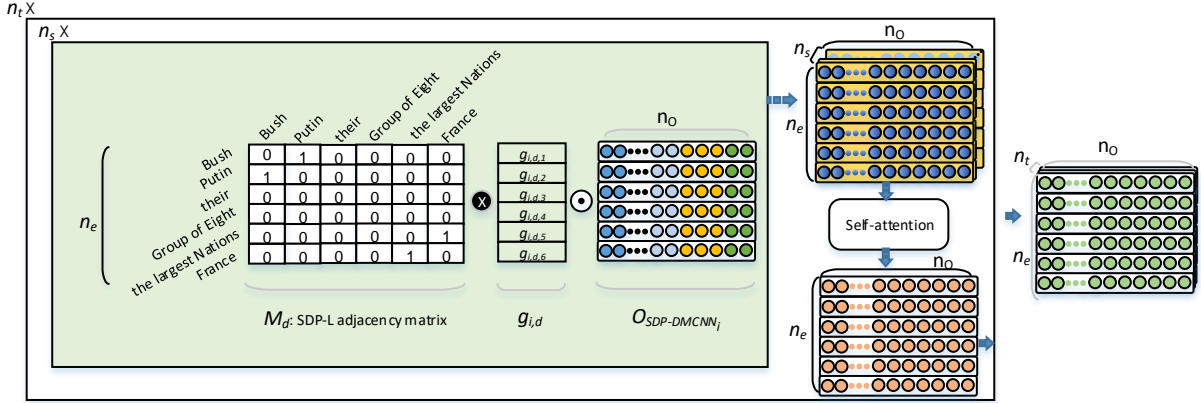

Figure 4: Illustration of the GCN-SDP architecture. The SDP-L adjacency matrix ($M_d$) is calculated for d=1.

The inputs of the GCN-SDP includes the SDP-L adjacency matrices and the output of the SDP-DMCNN module for trigger candidate $t_i$ ($O_{SDP-DMCNN_i} \in R^{n_e \times n_O}$) obtained from the previous step, where $1 \leq i \leq n_t$, $n_e$ is the number of argument candidates and $n_O$ is the length of SDP-DMCNN output vector. The ⊙ is the element-wise product operation. The graph convolution network for the trigger candidate $t_i$ and $d \leq n_s$ ($I_{i,d} \in R^{n_e \times n_O}$) is calculated by:

$$I_{i.d} = M_d (g_{i.d} \odot O_{SDP-DMCNN_i}) \tag{8}$$

where $M_d \in R^{n_e \times n_e}$ indicates the adjacency matrix for the SDP-L=d. Not all argument candidates are equally informative in interactions. Thus, $g_i \in R^{n_s \times n_e}$ is presented as a gate to weight the importance of argument candidates in different SDP-Ls. $g_{i.d} \in R^{n_e}$ for the trigger candidate $t_i$ and $d \leq n_s$ is calculated as follows:

$$g_{i.d} = f(O_{SDP-DMCNN_i} W_d + b_d) \tag{9}$$

where $f$ is a sigmoid function, $W_d \in R^{n_o}$ and $b_d \in R$ are the weight matrix and the bias of the gate, respectively.

After stacking $n_s$ matrices which present the interaction information between argument candidates in different SDP-Ls, we use a self-attention layer to consider the importance of each SDP-L in the aggregation phase. We calculate the score of each SDP-L vector for the trigger candidate $t_i$ ($Score_{i.d} \in R^{n_e}$) and then the output of the GCN-SDP for the trigger candidate $t_i$ ($I_{GCN-SDP_i} \in R^{n_e \times n_o}$) is calculated by:

$$Score_{i.d} = norm\ (\exp(z_d(f(W_3 I_{i.d} + b_3))))$$
$$I_{GCN-SDP_i} = \sum_{d=0}^{n_s} Score_{i.d} \odot I_{i.d} \tag{10}$$

where the "*norm*" function is a normalization operation, $f$ is a *tanh* function, $W_3 \in R^{n_o}$ and $b_3 \in R$ are the weight matrix and the bias of the gate, respectively. $z \in R^{n_s}$ is a vector which considers the importance of each SDP-L in presenting relationships between argument candidates. In order to simplify the explanation, we consider one trigger candidate in Equations 8-10. Thus, these equations are calculated for each trigger candidate $t_i$ in the sentence and then, their outputs are stacked together.

### 3.2.3 Argument Identification or Role Classification Output

Finally, to predict the roles of argument candidates for each trigger candidate, the outputs of the GCN-SDP ($I_{GCN-SDP} \in R^{n_t \times n_e \times n_o}$) and the SDP-DMCNN ($O_{SDP-DMCNN} \in R^{n_t \times n_e \times n_o}$) are concatenated and feed into a fully-connected network as follows:

$$O_{ij} = f(W_4[O_{SDP-DMCNN_{ij}} \oplus I_{GCN-SDP_{ij}}] + b_4)$$
$$y_{ij} = softmax(W_5 O_{ij} + b_5) \tag{11}$$

where, $1 \leqslant i \leqslant n_t$, $1 \leqslant j \leqslant n_e$ and $f$ is a non-linear activation, and $y_{ij} \in R^{n_r}$ calculates a confidence score according to the role of *j-th* argument candidate for *i-th* trigger candidate. Also, $n_r$ is the number of roles in the argument role classification task with "*NoRole*" label and is equal to 2 in the argument identification task (it shows that the argument candidate is an argument for a particular event trigger or not).

## 4 Experiments

In this section, we first introduce the dataset and evaluation metrics (Section 4.1). In the next section, the training phase and hyperparameters of *JEE-SDP* framework are presented (Section 4.2). Then, we evaluate our framework on the ACE dataset and compare it with the baseline methods (Section 4.3). Finally, we illustrate the effectiveness of the different layers of our model (Section 4.4-4.6).

### 4.1 Dataset and Evaluation Metrics

In this work, we use the ACE dataset[24] as a means to evaluate the proposed framework. We rely on the fact that ACE is currently the most popular and the most cited dataset for the EE task. In addition, it has also been very useful in the named entity recognition (NER) and relation extraction tasks. To comply with the previous works [27, 29, 30, 39, 68], a pre-defined split of the documents is used resulting in a training set with 529 documents (14,840 sentences), a validation set with 30 documents (863 sentences), and a test set with 40 documents (672 sentences).

The ACE 2005 dataset includes 33 sub-types such as "*Marry*", "*Transport*", "*Attack*" and etc. grouped into 8 event types and 35 argument roles such as "*Buyer*", "*Artifact*", "*Victim*" and etc. The goal of the trigger classification task is to classify each token of a given sentence into 67 categories with the "*NoEvent*" in BIO annotation schema. Also, the goal of the argument role classification task is to classify the role of each argument candidate for each trigger candidate into 36 categories with the "*NoRole*". When a trigger / argument candidate is selected as the "*NoEvent*" / "*NoRole*" during classification, it means it is not identified.

---
[24] https://catalog.ldc.upenn.edu/ ldc2006t06

Similar to Liu et al. [29], the maximum length *L*=50 is considered by padding shorter sentences and cutting off longer ones in the experiments. We use the Spacy toolkit[25] for tokenizing, POS-tagging and dependency parsing on the ACE dataset.

Same as in the previous works [27, 29, 30, 39, 68], we also use the following criteria to judge the correctness of the event extraction output. A trigger is correctly classified if its event sub-type and offsets match those of a reference trigger. An argument is correctly identified if its event sub-type and offsets match those of any of the reference argument mentions. An argument is correctly classified if its event sub-type, offsets and argument role match those of any of the reference argument mentions. To this regard we use *Precision (P)*, *Recall (R)* and *F-measure (F1)* as the evaluation metrics.

**4.2 Training and Hyperparameter Setting**

*JEE-SDP* framework is implemented as a multi-output model by Keras Deep Learning library[26]. During the validation and testing phases, the maximum score is selected as an event trigger label and argument role. Table 1, shows the hyperparameters of *JEE-SDP* tuned on the validation set through grid search.

Table 1: Hyperparameters in *JEE-SDP*

| **Hyperparametes** | |
|---|---|
| Glove embedding size | 300 |
| Bert embedding size | 768 |
| Dependency relation embedding size | 15 |
| Argument type embedding size | 10 |
| POS-tag embedding size | 15 |
| Positional embedding size (PF) | 10 |
| LSTM hidden units | 64 |
| Regularization | $10^{-4}$ |
| Dropout fraction | 0.5 |
| Batch size | 32 |
| The number of filters in DMCNN | 50 |
| The convolution window size in DMCNN | 3 |
| Hidden unit number (in the trigger classification output) | 256 |
| Hidden unit number (in the argument identification output) | 256 |
| Hidden unit number (in the argument role classification output) | 512 |

The network parameters are $\theta = [\overleftarrow{LSTM}, \overrightarrow{LSTM}, W_1, W_2, b_1, b_2, W_j, b_j, Z_d, W_d, b_d, W_3, W_4, W_5, b_3, b_4, b_5]$ in *JEE-SDP* where $0 \leqslant d \leqslant n_s$ in the GCN-SDP and $1 \leqslant j \leqslant n_k$ in DMCNN. We used ReLU [69] as our nonlinear activation function in *JEE-SDP*. To training these parameters, the cross entropy is selected as loss function with equal weight for two output models. We use the stochastic gradient descent algorithm with shuffled mini-batches and Adam update rule [70] to minimize the loss functions.

---

[25] https://spacy.io/
[26] https://keras.io/

## 4.3 Results of Event Extraction

The EE task on the ACE dataset began in 2004 and many researchers have been working on it. In this section, we compare the effectiveness of our framework with the strong baselines considered in the evaluation part of our work. We divide them into feature-based [27, 53, 54], deep neural network-based [28-30, 39, 71] and data augmentation based-methods [12, 34]. Following, the state-off-the-art methods in event extraction task are presented into three types:

- **Feature-based methods:**
    - *Cross-event* method [53] extracts features in document-level based on the co-occurrence between events and arguments.
    - *Cross-entity* method [54] extracts features according to relations between entities.
    - *JointBeam* method [27] proposes a structure-based system by manually designed global features which explicitly capture the dependencies of multiple triggers and arguments.
- **Deep neural networks-based methods:**
    - *DMCNN* method [39] uses dynamic multi-pooling to extract the best features from the different parts of a sentence according to the position of trigger and argument candidate.
    - *JRNN* method [28] proposes a joint framework with bidirectional recurrent neural networks and manually designed features to jointly extract event triggers and arguments.
    - *RBPB* method [71] uses a regularization method to consider two kinds of positive and negative argument relationships. RBPB uses trigger embedding, sentence-level embedding and pattern embedding together to improve the effectiveness of EE.
    - *DbRNN* method [30] models dependency relations between words via adds dependency bridges over Bi-LSTM. A tensor layer is used to simultaneously capture latent interaction between argument candidates.
    - *JMEE* method [29] models dependency relations between words by Graph Convolutional Networks (GCNs) to exploit syntactic information. The GCN layer enriches word representation by considering syntactic arcs in dependency parsing graph. In this work, we use the GCN layer to capture and aggregate the associations between argument candidates by defining the length of the SDP.
- **Data augmentation-based methods:**
    - *PLMEE* method [34] uses BERT pre-trained language model to generate labeled data due to insufficient training data by argument replacement and adjunct token rewriting.
    - *MLN-joint* method [12] proposes a joint event extraction approach based on Markov Logic Network (MLN), which includes the local information such as POS-tags and dependency relations and the global information extracted from frames expressing event information in FrameNet.

*JEE-SDP* is a deep neural networks-based method which is compared with the state-off-the-art methods in Table 2 on the blind test set. In Table 2, two versions of *JEE-SDP* are proposed, *JEE-SDP* without the GCN-SDP and *JEE-SDP* with the GCN-SDP via the self-attention layer ("*JEE-SDP +GCN-SDP +ATT*"). From the results, we can see that the *PLMEE* in the trigger classification task and "*JEE-SDP +GCN-SDP +ATT*" in the argument identification task and argument role classification task achieves the best effectiveness among all the competing methods according to F1-score. Regardless of the data augmentation methods, *JEE-SDP* improves F1-score by at least 2.1% for the trigger classification task compared to the deep neural networks-based methods. Also, "*JEE-SDP +GCN-SDP +ATT*" increases the F1-score by at least 6.1% compared to the deep neural networks-based methods and also, 3.69% over all the compared methods in the argument role classification task. This demonstrates the effectiveness of the proposed framework in EE. In the following, the effect of the different layers that are part of JEE-SDP framework are discussed.

Table 2: Overall effectiveness with gold-standard entities. Bold denotes the best result. The results of different methods are quoted from the corresponding papers due to the same test set. * shows the methods of our proposed framework.

| Type | Method | Trigger Classification (%) | | | Argument Identification (%) | | | Argument Role Classification (%) | | |
|---|---|---|---|---|---|---|---|---|---|---|
| | | P | R | F1 | P | R | F1 | P | R | F1 |
| Feature-based | *Cross-Event* (2010) | 68.7 | 68.9 | 68.8 | 50.9 | 49.7 | 50.3 | 45.1 | 44.1 | 44.6 |
| | *Cross-Entity* (2011) | 72.9 | 64.3 | 68.3 | 53.4 | 52.9 | 53.1 | 51.6 | 45.5 | 48.3 |
| | *JointBeam* (2013) | 73.7 | 62.3 | 67.5 | 69.8 | 47.9 | 56.8 | 64.7 | 44.4 | 52.7 |
| Deep neural networks-based | *DMCNN* (2015) | 75.6 | 63.6 | 69.1 | 68.8 | 51.9 | 59.1 | 62.2 | 46.9 | 53.5 |
| | *JRNN* (2016) | 66.0 | 73.0 | 69.3 | 61.4 | 64.2 | 62.8 | 54.2 | 56.7 | 55.4 |
| | *RBPB* (2016) | 70.3 | 67.5 | 68.9 | 63.2 | 59.4 | 61.2 | 54.1 | 53.5 | 53.8 |
| | *DbRNN* (2018) | 74.1 | 69.8 | 71.9 | 71.3 | 64.5 | 67.7 | 66.2 | 52.8 | 58.7 |
| | *JMEE* (2018) | 76.3 | 71.3 | 73.7 | 71.4 | 65.6 | 68.4 | 66.8 | 54.9 | 60.3 |
| | *JEE-SDP** | | | | 75.38 | 65.92 | 70.33 | 65.72 | 62.15 | 63.89 |
| | *JEE-SDP +GCN-SDP +ATT** | 81.16 | 71.16 | 75.83 | 76.64 | **67.03** | **71.52** | 66.08 | **66.81** | **66.44** |
| Data augmentation-based | *PLMEE* (2019) | 81.0 | **80.4** | **80.7** | 71.4 | 60.1 | 65.3 | 62.3 | 54.2 | 58.0 |
| | *MLN-Joint* (2019) | **85.69** | 71.68 | 78.06 | **80.99** | 58.43 | 67.88 | **74.86** | 54.01 | 62.75 |

### 4.3 Effect of the Different Word Representation Methods

As mentioned in the embedding layer section, we used Bert and Glove embedding methods in *JEE-SDP* for the trigger classification task. Table 3 shows the results for the trigger classification task when considering different representations.

The results obtained show that Bert embedding outperforms Glove in the trigger classification task. One reason for this might be due to the fact that Bert method considers the context of words in the word representation layer. Also, using LSTM layers after the pre-trained BERT may have an impact in the representation of the semantic of words in a sentence.

We concluded that using Bert does not provide significant improvement in the argument role classification task. The effectiveness of the argument role classification task could be affected by layers which consider the relations between event trigger and argument candidates in *JEE-SDP*. However, concatenating Bert and Glove improves the F1-score by 2%, which means Glove and BERT maybe complementary in the trigger classification task. This demonstrates the effectiveness of the embedding methods in solving the **ambiguity in classifying event triggers**.

Table 3: Effect of the different word representation methods in *JEE-SDP*

| Stage | Representation | precision | recall | F1-score |
|---|---|---|---|---|
| Trigger classification | Glove | 73.79 | 67.44 | 70.47 |
| | Bert | 80.49 | 68.13 | 73.80 |
| | BERT + Glove | **81.16** | **71.16** | **75.83** |

### 4.4 Effect of Extracting Multiple Events

To evaluate the effect of *JEE-SDP* framework in sentences which include multiple events, the test set is divided into two parts (1/1 and 1/N) according to the number of event triggers in the sentences and the effectiveness is evaluated separately [28, 29, 39]. (1/1) includes sentences which only have one event trigger; otherwise, 1/N is used.

Table 4 illustrates the effectiveness of the EE methods on the single and multiple event triggers sentences. To consider the effect of each layer, we present the results of *JEE-SDP* framework with and without some layers separately. According to Table 4, since joint approaches such as *JRNN*, *JMEE* and *JEE-SDP* would result in capturing co-occurrence relationships between event trigger subtypes in sentences, they present a significant improvement in EE especially in sentences with multiple events (1/N). For example, where the model predicted an *Attack* event, another event trigger subtypes can be most likely *Die*, *Transport* and *Injure* [29].

Table 4: The effectiveness of EE on single event sentences (1/1) and multiple event sentences (1/N) based on F1-score. (This table includes the methods which mentioned the results of their experiment on (1/1) and (1/N) sentences.)

| Stage | Method | 1/1 | 1/N | all |
|---|---|---|---|---|
| Trigger classification | DMCNN (2015) | 74.3 | 50.9 | 69.1 |
| | JRNN (2016) | 75.6 | 64.8 | 69.3 |
| | JMEE (2018) | 75.2 | 72.7 | 73.7 |
| | JEE-SDP* | **81.69** | **74.55** | **75.83** |
| Argument role classification | DMCNN (2015) | 54.6 | 48.7 | 53.5 |
| | JRNN (2016) | 50.0 | 55.2 | 55.4 |
| | JMEE (2018) | 59.3 | 57.6 | 60.3 |
| | JEE* | 55.61 | 60.26 | 58.51 |
| | JEE-SDP* | 59.73 | 66.48 | 63.89 |
| | JEE-SDP +TNT* | **63.09** | 65.63 | 64.65 |
| | JEE-SDP +GCN-SDP −ATT* | 60.53 | 67.71 | 64.95 |
| | JEE-SDP +GCN-SDP +ATT* | 61.40 | **69.69** | **66.44** |

*JEE-SDP* improves F1-score by at least 6.1% and 1.9% for the trigger classification task in (1/1) and (1/N) sentences, respectively. Also, the most important observation from the table is that "*JEE-SDP +GCN-SDP +ATT*" significantly outperforms all the other methods with large margins especially in (1/N) sentences for the argument role classification task which yields 12.1% improvement compared to the state-of-the-art method (*JMEE* method). This demonstrates that the proposed framework can effectively capture more valuable clues when a sentence contains more than one event.

Since the SDP extraction and the GCN-SDP focused on relations between event triggers and argument candidates, they could not significantly affect the effectiveness of the trigger classification task. Thus, the effectiveness of these layers is evaluated only in the argument role classification task in the following sub-sections.

### 4.5 Effect of the SDP Extraction Layer

To consider the effectiveness of the SDP extraction layer in our framework, according to table 4, we proposed *JEE* method as a joint approach via a DMCNN layer without considering the SDP extraction layer and the GCN-SDP which achieves F1-score by 58.51% in the argument role classification task.

In Table 4, *JEE-SDP* includes *JEE* method via the SDP extraction layer which obtains a substantial improvement by 5.38% compared to *JEE* and 3.6% compared to *JMEE* in all sentences. Also, using the SDP extraction layer improve the effectiveness of the argument role classification task in (1/1) and (1/N) sentences by 4.12% and 6.22% compared to *JEE*, respectively. In Figure 5, we show the effectiveness of the two methods *JEE* and *JEE-SDP* in the argument role classification task according to the sequential distance between event triggers and their arguments on the test set. According to this figure, the SDP extraction layer could successfully eliminate irrelevant words to help in extracting valuable features both in the short and the long range dependences between event triggers and their arguments in the sentences. The best effectiveness of *JEE-SDP* belongs to the bin 1-3 (the sequential distance between event trigger and argument candidate is between 1 to 3) which achieves F1-score by 86% in the argument role classification task.

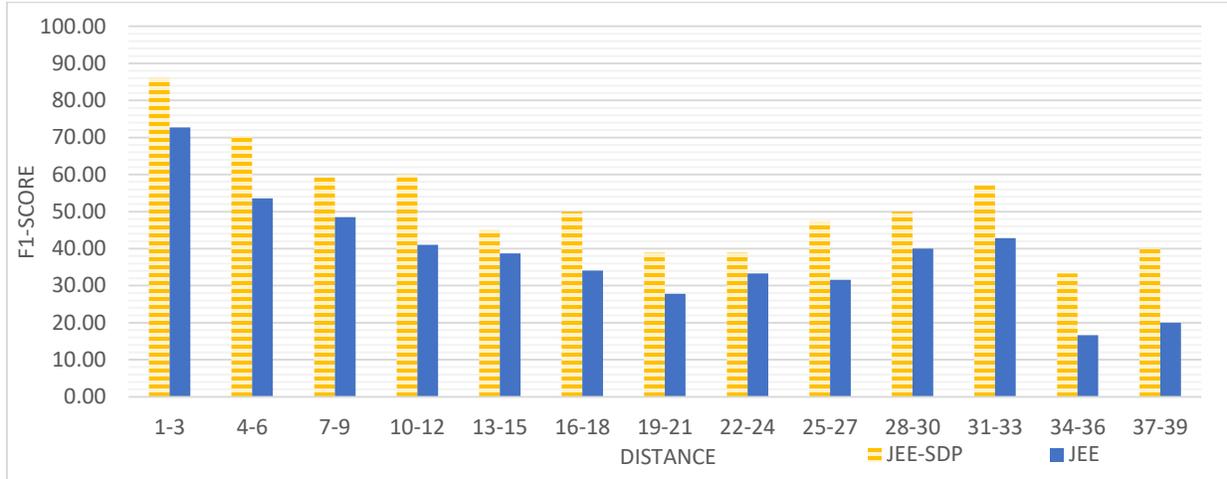

Figure 5: The effectiveness of two models *JEE* and *JEE-SDP* based on the sequential distance between event triggers and their arguments. The sequential distance is divided into bins of width 3.

**4.6 Effect of the GCN-SDP + the Self-Attention Layer**

To consider the effectiveness of layers which focus on extracting associations between arguments candidates, the results of *JEE-SDP* are presented via three methods *"JEE-SDP +TNT"*, *"JEE-SDP +GCN-SDP -ATT"* and *"JEE-SDP +GCN-SDP +ATT"* in Table 4. As mentioned previously, the TNT[27] layer is applied in EE to model interactions between argument candidates [26, 30]. According to this table, *"JEE-SDP +TNT"* improved the effectiveness of the argument role classification in (1/1) and all sentences by 3.36% and 0.76% compared to *JEE-SDP*. Also, this method has the best effectiveness in (1/1) sentences by 1.69% compared to *"JEE-SDP +GCN-SDP +ATT"*. However, TNT layer reduced the effectiveness in (1/N) sentences by 0.85% compared to *JEE-SDP*. It means that, the interaction features extracted by tensor layer are useful when each argument plays at most one role in the sentence, while, the tensor layer could not model complex relations between argument candidates in (1/N) sentences.

The GCN-SDP without a self-attention layer (*"JEE-SDP +GCN-SDP –ATT"*) improved the effectiveness of *JEE-SDP* by 0.8%, 1.23% and 1.06% in (1/1), (1/N) and all sentences, respectively. In the GCN-SDP without a self-attention layer, after stacking the different SDP-L vectors, we simply sum them to get a vector for each argument candidate. Thus, the importance of different SDP-Ls is not considered. To solve this problem, we used a self-attention layer to learn weights for different SDP-Ls.

*"JEE-SDP +GCN-SDP +ATT"* outperforms *JEE-SDP* by 1.67%, 3.21% and 2.55% in (1/1), (1/N) and all sentences, respectively. Also, using the self-attention layer improves the effectiveness of *"JEE-SDP +GCN-SDP +ATT"* by 1.49% compared to *"JEE-SDP +GCN-SDP -ATT"* in all sentences. This demonstrates that the GCN-SDP via the self-attention layer is useful in capturing the possible inter-dependencies extracted among argument candidates, especially when a sentence contains multiple events. Figure 6 shows the weight of each SDP-L in the aggregation phase which is learned in the self-attention layer by the parameter $z$. The higher the weight, the more impact on the aggregation phase.

---

[27] Tensor neural tensor

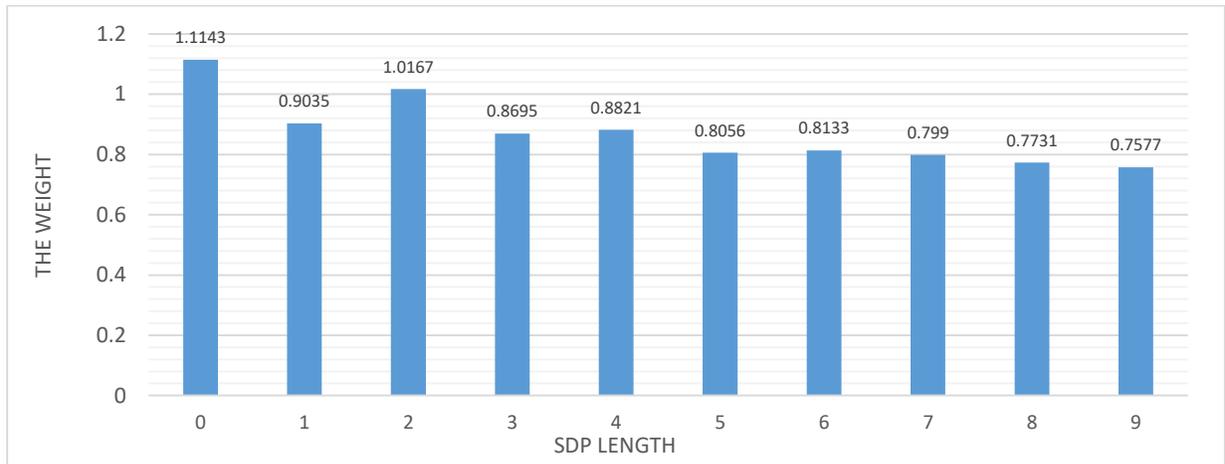

Figure 6: The visualization of trained weights for different SDP lengths ($z_d$).

As we mentioned previously, associations between argument candidates are calculated according to the SDP-L between them. The SDP-L=0 demonstrates the impact each argument candidate with itself in aggregation phase, which has the highest weight. The second impact belongs to the SDP-L=2, which typically captures the relations between argument candidates which have the same parent in the dependency graph. The third impact belongs to the SDP-L =1, which typically presents the relations between argument candidates paralleled in a sentence. It shows that the self-attention layer successfully models the inter-dependencies between argument candidates. Other weights for different SDP-Ls are typically decreased when the length of SDPs between arguments are increased.

## 5 Conclusion

In this paper, we propose *JEE-SDP* framework which models syntactic information by defining several neural network layers for the event extraction task. This framework predicts event triggers and their arguments simultaneously in a sentence. In the trigger classification task, we use BERT and Glove embedding methods together via a Bi-LSTM layer to consider the context of words to tackle the **ambiguity problem in classifying event triggers.** We demonstrate that our framework gains at least 2.1% improvement in terms of F1-score compared to the deep neural networks state-of-the-art methods.

In the argument role classification task, we apply the shortest dependent path (SDP) between event trigger and argument candidate in the dependency graph to solve **very long-range dependencies problem** by eliminating irrelevant words in the sentence. Using the SDP extraction layer improved the effectiveness of our framework by 5.38% F1-score in the argument role classification task.

Also, we use graph convolution network (GCN) via the self-attention layer to capture and aggregate the associations between argument candidates when nodes are argument candidates and the label of edges is the length of SDP (SDP-L). We showed that using the SDP-L between argument candidates significantly captures syntactic information such as having the same parent and parallel arguments in a sentence. We also showed that using the self-attention layer helps *JEE-SDP* to learn weight of each SDP-L in aggregation phase. Using GCN via the self-attention layer improved the effectiveness of our framework by 2.55% F1-score in the argument role classification task. Also, *JEE-SDP* framework outperforms all of the other methods by at least 3.69% F1-score in the argument role classification task. In the future, we plan to consider the association between event triggers to enhance the prediction of their subtypes in the trigger classification task.

## Acknowledgements

The third author of this paper, Ricardo Campos, was financed by the project PTDC/CCI-COM/31857/2017 (NORTE-01-0145-FEDER-03185) and also by the Portuguese funding agency, FCT - Fundação para a Ciência e a Tecnologia, through national funds, and co-funded by the FEDER, where applicable.